\documentclass[10pt,twocolumn,letterpaper]{article}

\usepackage[pagenumbers]{cvpr} %

\usepackage[table]{xcolor}
\usepackage{graphicx}
\usepackage{amsmath,amssymb,amsfonts}
\usepackage{booktabs}
\usepackage{multirow}
\usepackage{tabularx}
\usepackage{makecell}
\usepackage{bm}
\usepackage{nicefrac}
\usepackage{microtype}
\definecolor{cvprblue}{rgb}{0.21,0.49,0.74}
\usepackage[pagebackref,breaklinks,colorlinks,allcolors=cvprblue]{hyperref}

\definecolor{rankA}{HTML}{A9C8E8}   %
\definecolor{rankB}{HTML}{C8DEF2}   %
\definecolor{rankC}{HTML}{E4EFFA}   %
\newcommand{\rankone}[1]{\cellcolor{rankA}\textbf{#1}}
\newcommand{\ranktwo}[1]{\cellcolor{rankB}#1}
\newcommand{\rankthree}[1]{\cellcolor{rankC}#1}

\newif\ifrevmark \revmarkfalse 
\newcommand{\rev}[1]{\ifrevmark{\color{blue}#1}\else#1\fi}

\def\confName{CVPR}
\def\confYear{2027}

\title{GeoProto: Geometry-aware Prototype Learning for Cross-domain Few-shot Medical Image Segmentation}

\author{
Feifan Song$^{1}$ \qquad
Yuntian Bo$^{2}$ \qquad
Xiatian Zhu$^{1}$ \qquad
Haofeng Zhang$^{2}$ \\[1ex]
$^{1}$CVSSP, University of Surrey \qquad
$^{2}$Nanjing University of Science and Technology \\[0.5ex]
{\tt\small f.song@surrey.ac.uk, yuntian.bo@njust.edu.cn, xiatian.zhu@surrey.ac.uk, zhanghf@njust.edu.cn}
}

\begin{document}
\maketitle

\begin{abstract}
  Cross-domain few-shot medical image segmentation (CD-FSMIS) requires a model to generalise simultaneously to novel anatomical categories and unseen imaging domains from only a handful of annotated examples. 
  Rather than suppressing domain-specific signals as most existing approaches do, we present a new perspective that exploits human anatomical structure as a stable reference for reliable matching under domain shift: despite drastic appearance changes across domains, the interior geometry of each organ, from boundary to centroid, remains consistent across individuals.
  In our proposed GeoProto,
  the core component, Geometry-Aware Prototype Enrichment (GAPE), augments each local appearance prototype with a learned geometric offset encoding its ordinal position within the organ's interior, so that regions with similar appearance but distinct anatomical strata become separable. These strata are predicted by an auxiliary Ordinal Shape Branch (OSB) trained under an ordinally consistent objective that enforces smooth, monotonic variation across adjacent interior strata, requiring no annotation beyond standard binary masks.
  Extensive experiments across seven datasets spanning three evaluation settings demonstrate that GeoProto achieves state-of-the-art performance.
  Code will be released.

\end{abstract}

\section{Introduction}
\label{sec:intro}

\begin{figure}[t]
  \centering
  \includegraphics[width=0.95\linewidth]{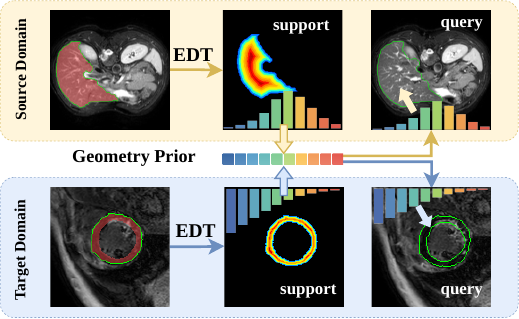}
  \caption{\small{Motivation of GeoProto. Despite substantial appearance differences across imaging domains, the anatomical structure of each organ, encoded as ordinal strata from boundary to centroid via EDT, remains consistent within each domain across individuals, providing a reliable, domain-transferable prior for cross-domain prototype matching.}}
  \label{fig:motivation}
  \vspace{-2ex}
\end{figure}

Few-shot medical image segmentation (FSMIS)~\cite{Ouyang2020proto, Tang2021RecurrentMR} alleviates the scarcity of pixel-level annotations~\cite{Ronneberger2015UNetCN, Isensee2020nnUNetAS} in medical imaging by segmenting novel categories from only a handful of labelled examples. In clinical practice, multiple imaging modalities (\eg, CT, MRI) are routinely employed to capture holistic diagnostic information, which then motivates the cross-domain FSMIS (CD-FSMIS) task~\cite{Zhu2024RobustEMDDR}.

Prototypical networks offer an efficient approach for FSMIS, where prototypes aggregated from support features are matched against the query~\cite{Ouyang2022SelfSupervisedLF, Bo2025FAMNetFM, Bo2025ContrastiveGM}.
In the cross-domain scenario, the model is trained on the source domain and applied directly to the unseen target domain, which is practical yet challenging.
Most existing CD-FSMIS methods address this domain shift by suppressing domain-specific signals, at the cost of discarding critical organ details. 
Rather than suppression, we argue that reliable cross-domain generalisation relies on a stable, domain-invariant reference.

Such a reference is, in fact, naturally available in medical imaging. Unlike natural images, where object shape and scale are highly variable, human anatomy exhibits a remarkable geometric regularity: \textit{despite substantial appearance variation across imaging modalities, the geometric structure of anatomical organs remains inherently consistent across individuals, deforming substantially only under pathological conditions.} We formalise this regularity as the spatial relationship between organ boundaries and their interior, which we term \emph{\textbf{anatomical strata}}\footnote{Inspired by \cite{Wang2019DDT}, we define the strata as concentric layers of an organ, obtained by quantising its normalised Euclidean distance-to-boundary transform (EDT) into ordered bins (see Sec.~\ref{sec:sdfhead} and Supp.~\ref{app:bin_map}).}. 

This regularity constitutes a simple, stable, and domain-transferable intra-organ prior that existing methods \cite{Ouyang2022SelfSupervisedLF,Bo2025FAMNetFM, Bo2025ContrastiveGM,zhu2025cross} have overlooked. As illustrated in Fig.~\ref{fig:motivation}, despite substantial appearance differences across imaging domains, the anatomical strata of each organ remain consistent across individuals. Under the few-shot setting, we derive this geometric representation directly from the support samples, forming the structural scaffold that anchors prototype-to-query matching along the stable boundary-to-centroid axis. 

Based on the observation above, we propose GeoProto, a geometry-aware CD-FSMIS framework that enriches prototypical matching with explicit structural priors.
We design two modules to inject this prior. First is Geometry-Aware Prototype Enrichment (GAPE), which extracts local prototypes from the backbone and augments them with learned geometric offsets, derived from the predicted distance-to-boundary bin maps of the organ's interior.
As a result, local prototypes with similar appearance yet distinct anatomical strata are displaced to geometrically differentiated regions of the feature space. To provide reliable geometric supervision, we further introduce the Ordinal Shape Branch (OSB), trained jointly with the segmentation model. This branch employs an ordinally consistent objective that penalises bin errors in proportion to their ordinal gap, enforcing smooth, monotonic variation of the predicted strata from boundary to centroid. Crucially, our supervision is derived entirely from standard binary masks, requiring no additional annotation.

In summary, our main contributions are:
\begin{itemize}
  \setlength{\itemsep}{2pt}
  \setlength{\parskip}{2pt}
  \item We present a new perspective on CD-FSMIS: rather than suppressing domain-specific signals at the cost of critical organ details, we exploit anatomical geometry, which remains consistent across individuals and imaging domains, as a stable anchor for cross-domain generalisation.
  \item We propose GeoProto, which realises this idea through two modules: GAPE enriches local prototypes with geometric offsets for geometry-conditioned matching, 
  and OSB learns the underlying strata via an ordinal objective.
  Both are supervised solely by segmentation masks, requiring no additional annotation.
  \item We conduct experiments on seven benchmarks spanning cross-modality, cross-sequence, and cross-context settings, demonstrating state-of-the-art overall performance.
\end{itemize}

\section{Related Work}
\label{sec:related}

\noindent \textbf{Few-Shot Medical Image Segmentation.}
FSMIS \cite{Ouyang2020proto,Lin2023CATNet} addresses the scarcity of medical annotations by leveraging class-agnostic knowledge to segment novel categories from only a few labelled examples. Prototypical network-based approaches \cite{Zhu2023proto, Tang2024proto, Cheng2025proto, Ouyang2020proto, Zhang2024Proto} dominate FSMIS, representing each class as one or more prototypes aggregated from support features and matched against query features to identify target regions. One line of work focuses on improving prototype quality: graph-based reasoning \cite{Huang2024Proto-graph} propagates support information while preserving query context, and high-fidelity prototypes \cite{Tang2024proto} preserve both semantic and structural cues for improved matching accuracy. Another line improves generalisation through richer supervision: superpixel pseudo-labels \cite{Ouyang2022SelfSupervisedLF} diversify episodic training tasks, prior mask generation \cite{Cheng2025PriorMask} provides additional query guidance, and cross-slice context modeling \cite{Hansen2022AnomalyDF} exploits the volumetric structure of medical images. However, these methods are designed for single-domain settings and degrade substantially under modality shift \cite{Zhu2025MAUPTM}.

\vspace{1ex}
\noindent \textbf{Extending to Cross-domain Scenarios.} CD-FSMIS extends FSMIS to unseen imaging domains without retraining. Existing methods address domain shift by suppressing domain-specific cues or improving feature matching. RobustEMD~\cite{Zhu2024RobustEMDDR} suppresses texture-sensitive signals through Earth Mover's Distance to emphasise class-boundary consistency, while FAMNet~\cite{Bo2025FAMNetFM} matches support and query features by frequency band and down-weights domain-specific components. C-Graph~\cite{Bo2025ContrastiveGM} models inter-class relations with cross-domain prototype graphs, and DSM~\cite{zhu2025cross} re-weights support and query features and adaptively selects channels that are robust to domain shift.
Closest to ours, PSP~\cite{Xu2026PSP} observes that organ position and shape are consistent across modalities, encoding position as image-centred polar coordinates and shape as a global descriptor that re-weights the channels of a single prototype.
However, these priors are image- or organ-level summaries injected without geometric supervision, and none captures where within an organ each local feature lies, leaving local matching vulnerable to appearance-level domain shift.

\begin{figure*}[t]
  \centering
   \includegraphics[width=\linewidth]{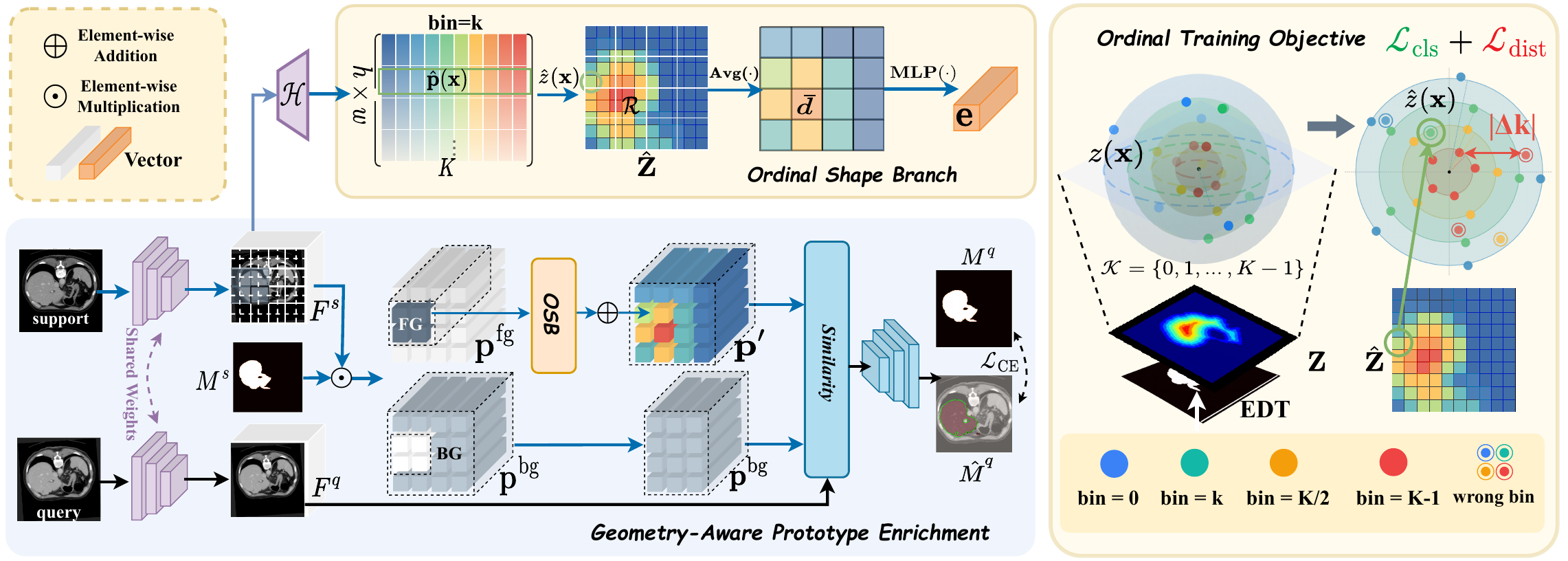}
          \vspace{-3ex}
  \caption{\small{
Overview of GeoProto. A shared backbone encodes support and query images. The Ordinal Shape Branch (OSB) decodes support features into a $K$-bin map of anatomical strata. It is supervised by an ordinal objective (right), which penalises errors in proportion to their bin gap $|\Delta k|$.
In Geometry-Aware Prototype Enrichment (GAPE), the expected stratum of each local foreground cell is mapped to a geometric offset $\mathbf{e}$ and added to its appearance prototype $\mathbf{p}^{\mathrm{fg}}$. The model is trained on the source domain and deployed directly on the unseen target domain without fine-tuning, and the pipeline is identical in both stages, except that all losses are removed at inference.}}
  \label{fig:overview}
  \vspace{-2ex}
\end{figure*}

\section{Methodology}
\label{sec:method}

\subsection{Problem Formulation}
\label{sec:prob}

Cross-domain few-shot medical image segmentation (CD-FSMIS) seeks to transfer a segmentation model $\Theta$, trained on a source domain $\mathcal{D}_s$ with labelled base categories $\mathcal{C}_b$, to an unseen target domain $\mathcal{D}_t$ containing novel categories $\mathcal{C}_n$, using only a handful of annotated examples. Crucially, $\mathcal{D}_s$ and $\mathcal{D}_t$ exhibit distributional shift induced by differences in imaging modality, acquisition protocol, or anatomical context, and their category sets are mutually exclusive, \ie, $\mathcal{C}_b \cap \mathcal{C}_n = \emptyset$.\footnote{Source-domain training uses only supervoxel pseudo-labels~\cite{Hansen2022AnomalyDF}.}
The model is deployed on $\mathcal{D}_t$ directly, without any retraining or fine-tuning.

Following the episodic meta-learning paradigm, we construct training tasks by randomly sampling episodes $(\mathcal{S}, \mathcal{Q})$ from $\mathcal{D}_s$. Each episode comprises a support set $\mathcal{S} = \{(I^s_i, M^s_i(c))\}_{i=1}^{S}$ of $S$ annotated image-mask pairs and a query set $\mathcal{Q} = \{(I^q, M^q(c))\}$, where $I$ denotes an input image slice, $M(c) \in \{0,1\}^{H \times W}$ its binary mask for the foreground class $c \in \mathcal{C}_b$, and $S$ is the number of support shots. At inference, the model is evaluated on episodes drawn from $\mathcal{D}_t$, wherein query images carry no annotation and the segmentation target $c$ is sampled from the novel set $\mathcal{C}_n$.
For brevity, we omit the shot index $i$ and class $c$ hereafter and write $M$ for a single mask.

\subsection{Overview of GeoProto}
\label{sec:overview}

Fig.~\ref{fig:overview} illustrates the GeoProto pipeline.
A shared ResNet \cite{He2015ResNet} backbone encodes both support and query images into features $F^s,  F^q\in \mathbb{R}^{C \times h \times w}$, where $C$ denotes the number of feature channels, and $h \times w$ is the feature spatial resolution.

In the Ordinal Shape Branch (OSB), support feature maps first pass through a lightweight decoder to produce per-pixel $K$-bin probabilities. These are then passed to the Geometry-Aware Prototype Enrichment (GAPE) module, which partitions support features into a $G{\times}G$ grid of non-overlapping cells, pools a masked-average prototype per foreground cell, and augments each prototype with a distance-aware geometric offset $\mathbf{e}$. Finally, the enriched prototypes $\mathbf{p}'$ and background prototypes $\mathbf{p}^{\mathrm{bg}}$ are matched against query features via softmax-weighted cosine similarity to produce the final prediction $\hat{M}^q$. The full training objective jointly optimises a segmentation loss, a bidirectional prototype alignment regulariser, and our geometry-aware OSB supervision.

\subsection{Ordinal Shape Branch}
\label{sec:sdfhead}

Medical organs exhibit predictable geometry: a consistent interior, smooth boundaries, and a stable centre-of-mass that generalises across domains. In this section, we discuss \textit{how to quantify this domain-agnostic geometry}.

\vspace{0.5ex}
\noindent \textbf{Motivation.} Our answer is a per-pixel geometric label that captures where each foreground pixel lies within the organ interior. A natural choice is the Euclidean distance-to-boundary transform (EDT), which assigns to each foreground pixel its shortest distance to the nearest background pixel. The EDT encodes the organ's interior topology in a continuous, geometry-faithful manner, which provides an ideal supervision signal for learning ordinal geometric representations.

\noindent \textbf{Bin construction.}
We uniformly partition the continuous distance-to-boundary range into $K$ discrete intervals (termed \textbf{bins}), such that each foreground pixel is assigned a bin index reflecting its ordinal stratum within the organ. Since directly regressing continuous distances is harder to optimise, we rephrase this geometric distance estimation as a bin classification problem.

Let $\tilde{M}^s \in \{0,1\}^{h \times w}$ denote $M^s$ downsampled to the
feature resolution, and let $\Omega = \{\mathbf{x} \mid \tilde{M}^s(\mathbf{x}) = 1\}$
be the set of foreground pixels. We first compute the Euclidean distance-to-boundary transform $\mathrm{EDT}(\tilde{M}^s)$ over organ pixels, normalise it by the maximum interior distance of each slice, and quantise it uniformly into $K$ bins, yielding the ground-truth bin map $\mathbf{Z} \in \mathcal{K}^{h \times w}$, where $\mathcal{K}=\left \{ 0,1,\dots,K-1 \right \} $ denotes the discrete geometric stratum space: bin $0$ marks the organ boundary, bin $K{-}1$ marks the region nearest the centroid, and intermediate bins encode concentric interior strata (see Supp.~\ref{app:bin_map} for details). 

During the forward pass, a lightweight convolutional decoder $\mathcal{H}$ maps support features to per-pixel bin logits $\mathcal{H}(F^s) \in \mathbb{R}^{K \times h \times w}$, from which the predicted bin map is obtained as:
\begin{equation}
  \hat{\mathbf{Z}} = \arg\max_{k}\, \mathrm{softmax}\bigl(\mathcal{H}(F^s)\bigr)_k \;\in\; \mathcal{K}^{h \times w}.
  \label{eq:sdfhead}
\end{equation}

\noindent \textbf{Ordinal training objective.} We design two complementary terms to supervise OSB in learning an ordinally smooth bin representation. 
Let $\hat{p}_{k}(\mathbf{x})$ denote the predicted softmax probabilities over bins derived from logits, $z(\mathbf{x})$ the ground-truth bin, $\hat{z}(\mathbf{x}) = \arg\max_{k} \hat{p}_k(\mathbf{x})$ the predicted bin, and $\beta = \tfrac{1}{2|\Omega|}$ a foreground normalisation factor. The first term is a standard cross-entropy restricted to foreground pixels, which penalises the classification error for each geometric stratum equally:
\begin{equation}
  \mathcal{L}_{\mathrm{cls}}
    = -\beta \sum_{\mathbf{x}\in \Omega}
      \sum_{k=0}^{K-1}
      \mathbf{1}(z(\mathbf{x}) = k)
    \log \hat{p}_{k}(\mathbf{x}),
  \label{eq:clsloss}
\end{equation}
where $\hat{p}_{k}(\mathbf{x})$ is the probability that pixel $\mathbf{x}$ belongs to bin $k$. The second term penalises misclassified pixels proportionally to their ordinal gap from the ground truth, suppressing over-confident predictions that land far from the true stratum more strongly than adjacent-bin confusions, rather than treating all misclassifications equally: 
\begin{equation}
  \mathcal{L}_{\mathrm{dist}}
    = -\beta \sum_{\mathbf{x}\in \Omega}
        \frac{|\hat{z}(\mathbf{x}) - z(\mathbf{x})|}{K}
        \log\bigl(1 - \max_{k}\,\hat{p}_{k}(\mathbf{x})\bigr).
  \label{eq:dist}
\end{equation}
As shown in Figure~\ref{fig:episode}, as training episodes increase, $\hat{z}(\mathbf{x})$ varies monotonically with distance from boundary to centroid in the geometry stratum space, rather than collapsing abruptly at a single bin.
Together, the combined OSB loss is:
\begin{equation}
  \mathcal{L}_{\mathrm{OSB}}
    = \mathcal{L}_{\mathrm{cls}}
    + \lambda_{\mathrm{dist}}\,\mathcal{L}_{\mathrm{dist}},
  \label{eq:sdfloss}
\end{equation}
where $\lambda_{\mathrm{dist}}$ balances the contribution of the ordinal penalty relative to the classification term.

\subsection{Geometry-Aware Prototype Enrichment}
\label{sec:gape}

Prototype-based few-shot methods aggregate support features into a compact representation of the target class. As a result, prototypes capture \emph{how} an organ appears (\eg, texture, intensity) but remain blind to \emph{where} within the organ each local feature originates.
Having established a reliable ordinal representation of organ geometry via OSB, we now describe how this signal is injected into the prototype feature space to remedy this limitation.

\vspace{0.5ex}
\noindent \textbf{Local prototype pooling.}
Rather than pooling a global prototype, which discards all spatial structure, we partition each support feature map $F^s \in \mathbb{R}^{C \times h \times w}$ into a $G{\times}G$ grid of non-overlapping cells, each covering $\frac{h}{G}{\times}\frac{w}{G}$ feature pixels, inspired by \cite{Ouyang2022SelfSupervisedLF}. Then, a masked-average prototype is pooled for each foreground cell:
\begin{equation}
\mathbf{p}_{i}^{\mathrm{fg}}
= \frac{\sum_{\mathbf{x} \in \mathcal{R}_i}
        F^s(\mathbf{x})\,\tilde{M}^s(\mathbf{x})}
       {\sum_{\mathbf{x} \in \mathcal{R}_i}
        \tilde{M}^s(\mathbf{x})},
\label{eq:rawproto}
\end{equation}
where $\mathcal{R}_i$ is the spatial extent of cell $i$. Cells whose average foreground occupancy falls below threshold $\tau$ are discarded, yielding $n_{\mathrm{fg}}$ active foreground prototypes per shot.

\vspace{0.5ex}
\noindent \textbf{Geometric offset embedding.}
For each retained cell $i$, we compute the expected ordinal strata under the OSB bin probability distribution $\hat{p}_{k}(\mathbf{x})$:
\begin{equation}
  \bar{d}_{i}
    = \frac{1}{|\mathcal{R}_i|}
      \sum_{\mathbf{x} \in \mathcal{R}_i}
      \sum_{k=0}^{K-1} k\cdot\hat{p}_{k}(\mathbf{x})
    \;\in\; [0,\,K{-}1].
  \label{eq:avgbin}
\end{equation}
Averaging over $\mathcal{R}_i$ yields a scalar $\bar{d}_{i}$ characterising the mean interior level of cell $i$: $\bar{d}_i \approx 0$ indicates a boundary-concentrated cell, while $\bar{d}_i \approx K{-}1$ indicates one near the centroid. 
Then, a lightweight two-layer MLP translates the scalar into a $C$-dimensional embedding $\mathbf{e}_{i}$,
where all weights are initialised near zero, ensuring geometric influence accrues gradually as the OSB learns increasingly ordinal representations.

\vspace{0.5ex}
\noindent \textbf{Prototype enrichment.}
Finally, the enriched prototype is obtained by element-wise addition of
the appearance prototype and its geometric embedding:
\begin{equation}
  \mathbf{p}_{i}' = \mathbf{p}_{i}^{\mathrm{fg}} + \mathbf{e}_{i}.
  \label{eq:enriched}
\end{equation}

Two prototypes with similar raw appearance yet distinct geometric strata receive different offsets $\mathbf{e}_i$, displacing them to different positions along the boundary-to-centroid axis in the feature space. This endows the feature space with an explicit geometric topology, where proximity reflects not only appearance similarity but also structural correspondence within the organ. Cosine similarity against these enriched prototypes thus anchors matching along this stable axis, rendering it robust to domain-specific appearance variations in novel imaging domains.

Background prototypes are computed without enrichment, as the geometric stratum is undefined for background pixels (as justified in Supp.~\ref{app:bg_sdf}).

\begin{table*}[t]
\centering
\caption{Quantitative results (DSC \%) under cross-modality scenarios.}
\vspace{-1ex}
\label{tab:table1}
\small
\addtolength{\tabcolsep}{-3pt}
\newcolumntype{Y}{>{\centering\arraybackslash}X}
\begin{tabularx}{\textwidth}{l *{5}{Y} @{\hspace{1em}} *{5}{Y}}
\toprule
\multirow{2}{*}{Method}
& \multicolumn{5}{c}{Abdominal CT $\rightarrow$ MRI}
& \multicolumn{5}{c}{Abdominal MRI $\rightarrow$ CT} \\
\cmidrule(lr{1em}){2-6} \cmidrule(lr){7-11}
& Liver & LK & RK & Spleen & Mean
& Liver & LK & RK & Spleen & Mean \\
\midrule
SSL-ALP~\cite{Ouyang2022SelfSupervisedLF}
& 70.74 & 55.49 & 67.43 & 58.39 & 63.01
& 71.38 & 34.48 & 32.32 & 51.67 & 47.46 \\
ADNet~\cite{Hansen2022AnomalyDF}
& 50.33 & 39.36 & 37.88 & 39.37 & 41.73
& 64.25 & 37.39 & 25.62 & 42.94 & 42.55 \\
RPT~\cite{Zhu2023FewShotMI}
& 49.22 & 42.45 & 47.14 & 48.84 & 46.91
& 65.87 & 40.07 & 35.97 & 51.22 & 48.28 \\
GMRD~\cite{Cheng2024FewShotMI}
& 63.15 & 61.79 & 67.69 & 56.89 & 62.38
& 66.12 & 57.38 & 56.37 & 54.56 & 58.61 \\
PATNet~\cite{Lei2022CrossDomainFS}
& 57.01 & 50.23 & 53.01 & 51.63 & 52.97
& \ranktwo{75.94} & 46.62 & 42.68 & 63.94 & 57.29 \\
PMNet~\cite{Chen2024PixelMN}
& 64.50 & 60.16 & 61.83 & 51.80 & 59.57
& 66.82 & 39.21 & 30.87 & 47.49 & 46.10 \\
IFA~\cite{Nie2024IFA}
& 48.81 & 45.79 & 51.46 & 51.42 & 49.37
& 50.05 & 36.45 & 32.69 & 43.08 & 40.57 \\
APM-M~\cite{Tong2024LightweightFM}
& \rankthree{70.85} & 55.41 & 58.68 & 53.11 & 59.51
& \rankthree{74.48} & 56.01 & 49.83 & 64.12 & 61.11 \\
RobustEMD~\cite{Zhu2024RobustEMDDR}
& 60.16 & 66.34 & 70.26 & 53.71 & 62.61
& 69.82 & 63.79 & 50.34 & 59.88 & 60.95 \\
FAMNet~\cite{Bo2025FAMNetFM}
& \rankone{73.01} & 57.28 & \rankthree{74.68} & 58.21 & 65.79
& 73.57 & 57.79 & 61.89 & \rankthree{65.78} & 64.75 \\
C-Graph~\cite{Bo2025ContrastiveGM}
& 68.28 & \rankthree{69.70} & 70.17 & \ranktwo{61.36} & \rankthree{67.38}
& 69.60 & \ranktwo{70.00} & \rankthree{63.95} & 65.23 & \ranktwo{67.20} \\
DSM~\cite{zhu2025cross}
& \ranktwo{72.94} & 61.59 & 69.52 & 59.00 & 65.76
& \rankone{77.69} & 56.60 & 56.45 & 59.63 & 62.59 \\
PSP~\cite{Xu2026PSP}
& 68.73 & \ranktwo{72.01} & \rankone{75.75} & \rankthree{60.21} & \ranktwo{69.17}
& 64.18 & \rankthree{67.31} & \ranktwo{69.56} & \ranktwo{66.87} & \rankthree{66.98} \\
\midrule
Ours
& 62.31 & \rankone{72.97} & \ranktwo{74.85} & \rankone{67.67} & \rankone{69.45}
& 69.65 & \rankone{71.31} & \rankone{69.65} & \rankone{67.98} & \rankone{69.65} \\
\bottomrule
\end{tabularx}
\end{table*}

\begin{table*}[t]
\centering
\caption{Quantitative results (DSC \%) under cross-sequence scenarios.}
\vspace{-1ex}
\label{tab:table2}
\small
\addtolength{\tabcolsep}{-3pt}
\newcolumntype{Y}{>{\centering\arraybackslash}X}
\begin{tabularx}{\textwidth}{l *{4}{Y} @{\hspace{1em}} *{4}{Y}}
\toprule
\multirow{2}{*}{Method}
& \multicolumn{4}{c}{Cardiac LGE $\rightarrow$ b-SSFP}
& \multicolumn{4}{c}{Cardiac b-SSFP $\rightarrow$ LGE} \\
\cmidrule(lr{1em}){2-5} \cmidrule(lr){6-9}
& LV-BP & LV-MYO & RV & Mean
& LV-BP & LV-MYO & RV & Mean \\
\midrule
SSL-ALP~\cite{Ouyang2022SelfSupervisedLF}
& 83.47 & 22.73 & 66.21 & 57.47
& 65.81 & 25.64 & 51.24 & 47.56 \\
ADNet~\cite{Hansen2022AnomalyDF}
& 58.75 & 36.94 & 51.37 & 49.02
& 40.36 & 37.22 & 43.66 & 40.41 \\
RPT~\cite{Zhu2023FewShotMI}
& 60.84 & 42.28 & 57.30 & 53.47
& 50.39 & 40.13 & 50.50 & 47.00 \\
GMRD~\cite{Cheng2024FewShotMI}
& 76.23 & 36.87 & 62.91 & 58.67
& 66.69 & 47.19 & 58.21 & 57.36 \\
PATNet~\cite{Lei2022CrossDomainFS}
& 65.35 & 50.63 & 68.34 & 61.44
& 66.82 & 53.64 & 59.74 & 60.06 \\
PMNet~\cite{Chen2024PixelMN}
& 73.46 & 32.11 & 68.70 & 58.09
& 57.14 & 30.13 & 60.12 & 49.13 \\
IFA~\cite{Nie2024IFA}
& 64.04 & 43.22 & 74.58 & 60.61
& 68.07 & 36.07 & 60.42 & 54.85 \\
APM-M~\cite{Tong2024LightweightFM}
& 68.91 & 45.74 & 61.78 & 58.81
& 57.72 & 42.37 & 52.83 & 50.97 \\
RobustEMD~\cite{Zhu2024RobustEMDDR}
& 75.32 & 51.32 & 72.86 & 66.50
& 73.19 & 50.02 & 60.29 & 61.16 \\
FAMNet~\cite{Bo2025FAMNetFM}
& \ranktwo{86.64} & 51.84 & \rankthree{76.26} & 71.58
& \rankone{77.37} & 52.05 & 54.75 & 61.39 \\
C-Graph~\cite{Bo2025ContrastiveGM}
& \rankone{87.61} & \rankthree{55.22} & \rankone{79.76} & \ranktwo{74.20}
& 68.46 & \ranktwo{56.38} & \rankthree{65.14} & \rankthree{63.33} \\
DSM~\cite{zhu2025cross}
& 85.27 & 50.74 & 73.20 & 69.74
& 71.27 & 53.62 & 63.65 & 62.85 \\
PSP~\cite{Xu2026PSP}
& 85.39 & \ranktwo{57.04} & 75.43 & \rankthree{72.62}
& \rankthree{74.88} & \rankthree{54.25} & \ranktwo{66.91} & \ranktwo{65.34} \\
\midrule
Ours
& \rankthree{85.45} & \rankone{60.79} & \ranktwo{76.46} & \rankone{74.23}
& \ranktwo{74.91} & \rankone{56.39} & \rankone{70.02} & \rankone{67.10} \\
\bottomrule
\end{tabularx}
\vspace{-1ex}
\end{table*}

\begin{table*}[t]
\centering
\caption{Quantitative results (DSC \%) under cross-context scenarios. All methods are trained on abdominal CT.}
\vspace{-1ex}
\label{tab:table7_extended}
\small
\addtolength{\tabcolsep}{-3pt}
\newcolumntype{Y}{>{\centering\arraybackslash}X}
\begin{tabularx}{\textwidth}{l *{2}{Y} @{\hspace{1em}} *{4}{Y} @{\hspace{1em}} *{4}{Y}}
\toprule
\multirow{2}{*}{Method}
& \multirow{2}{*}{CXR} & \multirow{2}{*}{ISIC}
& \multicolumn{4}{c}{Cardiac b-SSFP MRI}
& \multicolumn{4}{c}{MI-PMR} \\
\cmidrule(lr{1em}){4-7} \cmidrule(lr){8-11}
& & & BP & MYO & RV & Mean
& Bladder & CG & Rectum & Mean \\
\midrule
SSL-ALP~\cite{Ouyang2022SelfSupervisedLF}
& 71.66 & 39.16 & 61.18 & 27.14 & 48.45 & 45.59
& 54.29 & 27.76 & 19.92 & 33.99 \\
ADNet~\cite{Hansen2022AnomalyDF}
& 40.32 & 22.11 & 46.61 & 28.47 & 42.93 & 39.34
& 36.51 & 25.38 & 28.14 & 30.01 \\
RPT~\cite{Zhu2023FewShotMI}
& 58.48 & 35.44 & \rankthree{69.35} & 42.79 & 53.11 & 55.08
& \rankthree{66.17} & 38.84 & 26.94 & 43.98 \\
GMRD~\cite{Cheng2024FewShotMI}
& 52.31 & 39.66 & 41.45 & 28.36 & 38.59 & 36.13
& 36.08 & 22.33 & 24.61 & 27.67 \\
PATNet~\cite{Lei2022CrossDomainFS}
& 77.65 & 42.88 & \ranktwo{69.88} & \rankthree{46.09} & 52.97 & \rankthree{56.31}
& 56.5 & 46.98 & 38.22 & 47.23 \\
PMNet~\cite{Chen2024PixelMN}
& 75.65 & 37.28 & 62.42 & 21.34 & \rankthree{57.93} & 47.23
& 57.08 & \ranktwo{51.54} & \rankthree{44.84} & \ranktwo{51.15} \\
IFA~\cite{Nie2024IFA}
& 76.21 & 42.59 & 66.42 & 27.89 & 39.78 & 44.70
& 55.77 & 38.09 & 31.16 & 41.67 \\
APM-M~\cite{Tong2024LightweightFM}
& 71.49 & 39.85 & 44.30 & 31.17 & 45.18 & 40.22
& 43.04 & 31.26 & 41.16 & 38.49 \\
RobustEMD~\cite{Zhu2024RobustEMDDR}
& 63.99 & \rankthree{46.33} & 37.77 & 41.44 & 42.48 & 40.56
& 44.93 & 21.81 & 20.07 & 28.94 \\
FAMNet~\cite{Bo2025FAMNetFM}
& 67.02 & 36.45 & 58.44 & 33.00 & 51.34 & 47.59
& 42.78 & 27.06 & 29.44 & 33.09 \\
C-Graph~\cite{Bo2025ContrastiveGM}
& \rankthree{78.38} & \rankone{50.42} & 63.70 & \ranktwo{46.32} & \rankone{66.00} & \ranktwo{58.67}
& 52.55 & 48.44 & \rankone{49.30} & 50.10 \\
PSP~\cite{Xu2026PSP}
& \ranktwo{81.33} & 35.86 & 63.76 & 42.69 & 52.69 & 53.05
& \ranktwo{67.76} & \rankthree{49.43} & 34.49 & \rankthree{50.56} \\
\midrule
Ours
& \rankone{84.50} & \ranktwo{47.45} & \rankone{71.26} & \rankone{46.75} & \ranktwo{65.09} & \rankone{61.03}
& \rankone{76.43} & \rankone{59.44} & \ranktwo{47.83} & \rankone{61.24} \\
\bottomrule
\end{tabularx}
\vspace{-2ex}
\end{table*}

\subsection{Query Prediction}
\label{sec:classification}

With enriched foreground prototypes $\{\mathbf{p}_i'\}$ and standard background ones $\{\mathbf{p}_j^{\mathrm{bg}}\}$, each query pixel $\mathbf{x}$ receives a foreground score via softmax-weighted cosine similarity over all retained foreground prototypes: 
\begin{equation}
  S_{\mathrm{fg}}(\mathbf{x}) = \sum_{i} w_i(\mathbf{x})\cdot\cos\bigl (F^q(\mathbf{x}),\,\mathbf{p}_{i}'\bigr),
  \label{eq:score}
\end{equation}
where $w_i(\mathbf{x}) = \mathrm{softmax}_i \bigl(\alpha \cos(F^q(\mathbf{x}), \mathbf{p}_i')\bigr)$ are attention weights over foreground prototypes and $\alpha$ is a temperature scaling factor, highlighting the prototypes most relevant to each query pixel (\eg, a boundary pixel is not diluted by centroid prototypes). The background score $S_{\mathrm{bg}}(\mathbf{x})$ is computed analogously using $\{\mathbf{p}_j^{\mathrm{bg}}\}$, and the final prediction is obtained as $\hat{M}^q= \arg\max[S_{\mathrm{bg}},\, S_{\mathrm{fg}}]$.

To regularise the feature space, we add a bidirectional alignment loss  \cite{Wang2019PANetFI} by swapping support and query roles: raw query prototypes $\mathbf{p}_i^{q}$, pooled from the query feature map using the predicted mask $\hat{M}^q$, are used to segment the support image:
\begin{equation}
  \mathcal{L}_{\mathrm{align}} = \mathcal{L}_{\mathrm{CE}}(\hat{M}^s,\, M^s), 
  \label{eq:align}
\end{equation}
where $\hat{M}^s$ denotes the support prediction under query prototypes. Geometric enrichment is deliberately withheld here to avoid a circular optimisation dependency between GAPE and the alignment branch. Finally, the full training objective is:
\begin{equation}
  \mathcal{L}
    = \mathcal{L}_{\mathrm{CE}}(\hat{M}^q, M^q)
    + \mathcal{L}_{\mathrm{align}}
    + \lambda_{\mathrm{geo}}\,\mathcal{L}_{\mathrm{OSB}},
  \label{eq:total_full}
\end{equation}
where $\lambda_{\mathrm{geo}}$ balances geometric supervision against the primary segmentation objective.
\section{Experiments}
\label{sec:experiments}

\begin{figure}[t]
  \centering
  \includegraphics[width=\linewidth]{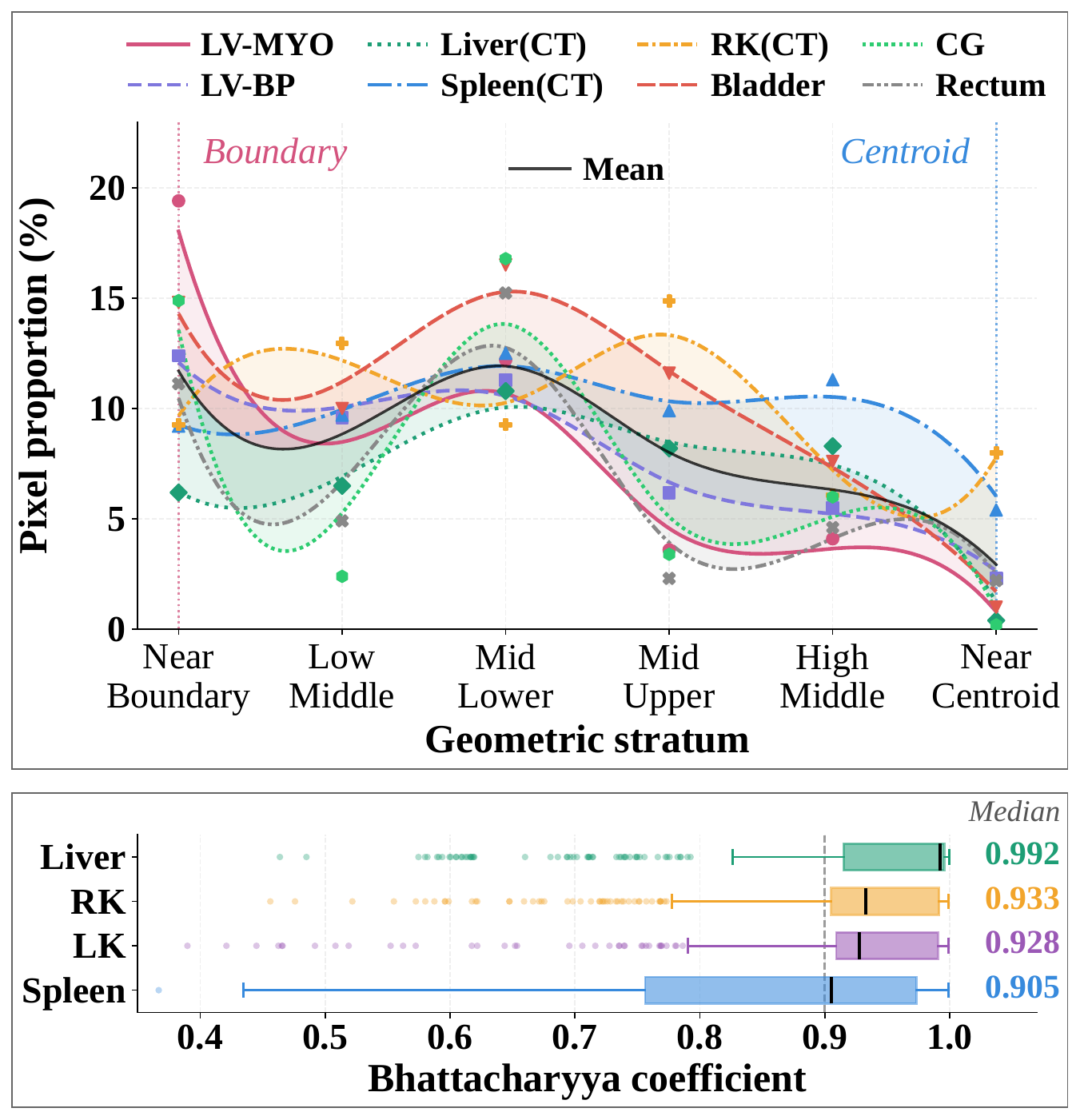}
  \vspace{-3ex}
  \caption{\small{(\textbf{Top}) EDT bin distributions across representative organs and datasets. (\textbf{Bottom}) Bhattacharyya coefficient (BC) between support and query EDT distributions over 500 intra-domain episodes per organ.}}
  \label{fig:geo_prior}
  \vspace{-4ex}
\end{figure}

\subsection{Experiment Setup}
\label{sec:experiment_setup}

\noindent {\bf Datasets.}
We evaluate on seven medical imaging datasets spanning four modalities.
\textbf{Abdominal CT} comprises 20 3D CT scans collected from the MICCAI 2015 Multi-Atlas Labeling Challenge~\cite{abd_ct} and \textbf{Abdominal MRI} comprises 20 3D T2-SPIR MRI scans from the ISBI 2019 Combined Healthy Abdominal Organ Segmentation Challenge~\cite{Kavur2020CHAOSC}.
\textbf{Cardiac b-SSFP} and \textbf{Cardiac LGE} are sourced from the MICCAI 2019
Multi-sequence Cardiac MR Segmentation Challenge  \cite{Zhuang2020CardiacSO, Zhuang2016MultivariateMM} with 45 3D cardiac MRI scans in each dataset acquired using
the b-SSFP and LGE sequences, respectively.
\textbf{MI-PMR} consists of 403 multi-institution prostate MRI scans sourced from UCLH  \cite{Dickinson2013MI-PRO} and NCI  \cite{choyke2016prostate} with annotations by  \cite{Li2022MI-PRO}. Three categories of labels are employed: Central Gland (CG), Bladder and Rectum. 
\textbf{Chest X-Ray} contains 704 X-Ray images for Tuberculosis released by the National Library of Medicine, Maryland, USA~\cite{Jaeger2014Lung, Candemir2014LungSI}.
\textbf{ISIC2018} provides 2594 dermoscopic images for skin lesion segmentation collected by  \cite{Codella2019SkinLA, Tschandl2018TheSkin}.

\vspace{0.5ex}
\noindent {\bf Experimental Protocol.}
We consider three evaluation settings: \textbf{(a) Cross-modality}: conducted on abdominal data, testing robustness to modality shift within the same anatomical region. 
\textbf{(b) Cross-sequence}: conducted on cardiac data acquired under different MRI sequences, isolating the effect of sequence variation on the same organ. 
\textbf{(c) Cross-context}: trained on Abdominal CT and evaluated on four structurally distinct targets, probing generalisation across anatomy, modality, and imaging context simultaneously. Throughout, we denote transfer directions as source\,$\to$\,target.
For a fair comparison, we adopt the Dice Similarity Coefficient (DSC)  \cite{Ouyang2020proto} as the evaluation metric.

\vspace{0.5ex}
\noindent {\bf Implementation Details.}
All experiments are implemented in PyTorch and trained on two NVIDIA RTX A6000 GPUs. As in  \cite{Hansen2022AnomalyDF}, supervoxel-based pseudo labels are generated to provide supervisory signals during model training. The backbone is a ResNet-101  \cite{He2015ResNet} initialised with MS-COCO  \cite{Lin2014MicrosoftCC} weights. We train for 40K episodes for cross-modality and cross-sequence settings, and 45K episodes for cross-context. All experiments follow a 1-way 1-shot protocol  \cite{Hansen2022AnomalyDF, Shen2022QNetQF, Zhu2023FewShotMI}. 

\vspace{0.5ex}
\noindent {\bf Hyperparameters.}
The prototype grid size is set to $G = 8$, \ie, the $32{\times}32$ feature map is partitioned into $8{\times}8$ cells of $4{\times}4$ feature pixels.
The OSB uses $K = 10$ distance-to-boundary bins. The training objective is weighted by $\lambda_{\mathrm{dist}} = 1.0$ for the ordinal penalty and $\lambda_{\mathrm{geo}} = 0.3$ for OSB supervision. The network is optimised with SGD ($\text{lr} = 10^{-3}$, momentum $= 0.9$, weight decay $= 5{\times}10^{-4}$), with the learning rate decayed by a factor of $0.95$ every 1k episodes.

\subsection{Comparison with State-of-the-Art}
\label{sec:sota}

We compare our method with SOTA methods, including CD-FSMIS methods RobustEMD \cite{Zhu2024RobustEMDDR}, FAMNet \cite{Bo2025FAMNetFM}, C-Graph \cite{Bo2025ContrastiveGM}, DSM~\cite{zhu2025cross},
PSP~\cite{Xu2026PSP}; CD-FSS methods PATNet \cite{Lei2022CrossDomainFS}, PMNet \cite{Chen2024PixelMN}, IFA \cite{Nie2024IFA}, APM-M \cite{Tong2024LightweightFM}; and FSMIS methods SSL-ALP \cite{Ouyang2022SelfSupervisedLF}, ADNet \cite{Hansen2022AnomalyDF}, RPT \cite{Zhu2023FewShotMI}, and GMRD \cite{Cheng2024FewShotMI}. 

As shown in Tables~\ref{tab:table1}--\ref{tab:table7_extended}, GeoProto consistently achieves the best overall performance across all three cross-domain settings. Under cross-modality transfer, GeoProto surpasses the strongest competitors PSP~\cite{Xu2026PSP} (69.17\%) on CT\,$\to$\,MRI and C-Graph~\cite{Bo2025ContrastiveGM} (67.20\%) on MRI\,$\to$\,CT, and performs exceptionally well on the kidneys and spleen. 
Under cross-sequence transfer, GeoProto outperforms the baseline SSL-ALP~\cite{Ouyang2022SelfSupervisedLF} by 18.15\% mean DSC on average, and PSP~\cite{Xu2026PSP} by 1.76\% on b-SSFP\,$\to$\,LGE.
Under the more challenging cross-context setting, GeoProto achieves the best results on CXR (84.50\%), Cardiac b-SSFP (61.03\%), and MI-PMR (61.24\%), with a 10.68\% gain over PSP~\cite{Xu2026PSP} on MI-PMR, while ranking second on ISIC. More statistical and sensitivity analyses can be seen in Supp.~\ref{app:statsens}.
These results suggest that our model acquires a genuine capacity for structural generalisation rather than domain-specific overfitting, effectively adapting to unseen anatomies and modalities from a single support sample.

\noindent \textbf{Compact and roughly convex structures.}
We also observe that the gains are most pronounced on compact, roughly convex organ structures that remain challenging for existing methods, \eg, the spleen (+6.31\% over the strongest competitor) under CT\,$\to$\,MRI, and the bladder (+8.67\%) and central gland (+7.90\%) on MI-PMR, where foreground pixels spread across multiple strata and a well-defined boundary-to-centroid structure provides a stable matching reference. Conversely, for large homogeneous organs such as the liver, 
most pixels collapse into the lowest strata, so appearance cues remain dominant (as substantiated in Supp.~\ref{app:result}).

\subsection{Ablation Study}
\label{sec:ablation}

\paragraph{Geometric prior validation.}
GeoProto rests on two premises: (\textit{i}) \textit{different organs possess distinct geometric stratum distributions}, and (\textit{ii}) \textit{these distributions remain consistent across individuals, enabling geometric priors to transfer from support to query}. 
As shown in Fig.~\ref{fig:geo_prior} (Top), thin-walled structures (\eg, LV-MYO) concentrate mass near the boundary strata, while compact organs (\eg, RK, Bladder) exhibit flatter distributions, empirically confirming that EDT bins carry discriminative structural information.
Fig.~\ref{fig:geo_prior} (Bottom) validates the second premise: all four abdominal organs achieve median BC~$>0.90$ (Liver: 0.992), confirming strong support--query geometric consistency. Spleen shows the greatest variability (BC~$=0.857\pm0.132$), attributable to its inter-subject shape irregularity.

\begin{table}[t]
\centering
\caption{Ablation study on geometric components, trained on Abdominal CT.
         ``PE'' denotes positional embedding, a control swapped in place of the geometric embedding;
         ``Geo-E'' denotes bin embedding, \ie, the geometric offset of Eq.~\ref{eq:enriched};
         ``OSB-L'' denotes the ordinal shape branch loss $\mathcal{L}_\text{OSB}$; ``w/o OSB-L'' keeps the OSB head but drops $\mathcal{L}_\text{OSB}$ from Eq.~\ref{eq:total_full}, so the head is trained only through the segmentation and alignment gradients.
         $\Delta$ denotes improvement over the baseline.}
\label{tab:ablation}
\setlength{\tabcolsep}{3pt}
\small
\begin{tabular}{l cc cc}
\toprule
\multirow{2}{*}{Method}
  & \multicolumn{2}{c}{Abdominal MRI}
  & \multicolumn{2}{c}{Cardiac b-SSFP} \\
\cmidrule(lr){2-3} \cmidrule(lr){4-5}
 & DSC (\%) & $\Delta$ & DSC (\%) & $\Delta$ \\
\midrule
Baseline
  & 63.01 & --
  & 45.59 & -- \\
\rowcolor{gray!12}  + PE
  & 66.37 & $+3.36$
  & 58.34 & $+12.75$ \\
+ Geo-E (w/o OSB-L)
  & 66.31 & $+3.30$
  & 54.88 & $+9.29$ \\
+ OSB-L (w/o Geo-E)
  & 68.13 & $+5.12$
  & 55.52 & $+9.93$\\
  \midrule
  Ours
  & \textbf{69.45} & $\textbf{+6.44}$
  & \textbf{61.03} & $\textbf{+15.44}$ \\
\bottomrule
\end{tabular}
\vspace{-2ex}
\end{table}

\paragraph{Component analysis.}
Table~\ref{tab:ablation} decomposes GeoProto against the baseline SSL-ALP \cite{Ouyang2022SelfSupervisedLF}, a widely used FSMIS framework, trained on Abdominal CT.
We first deploy positional embedding (PE) as an ablation control, which replaces the geometric embedding to verify the advantage over generic positional cues, yielding $+3.36\%$ on Abdominal MRI and $+12.75\%$ on Cardiac b-SSFP. Notably, Geo-E alone (without OSB-L) brings no further gain ($+3.30\%$, $+9.29\%$), suggesting that geometric offsets without ordinal supervision behave no better than PE, as the model lacks the structural signal needed to differentiate anatomical strata. Adding OSB-L alone surpasses PE on Abdominal MRI ($+5.12\%$) but not on Cardiac b-SSFP ($+9.93\%$), implying that ordinal supervision alone shapes the backbone yet is not sufficient. The full model combining both components achieves the best performance ($+6.44\%$, $+15.44\%$), demonstrating that Geo-E and OSB-L are complementary: ordinal supervision provides the structural grounding that makes geometric embeddings semantically meaningful rather than spatially redundant.

\begin{table}[t]
\centering
\caption{Comparison of computational complexity and inference efficiency. Latency is reported on a single RTX A6000 GPU.}
\label{tab:efficiency}
\small
\setlength{\tabcolsep}{2pt}
\begin{tabular}{l r r r}
\toprule
\multirow{2}{*}{Method}
  & \multicolumn{2}{c}{Latency (ms)}
  & \multirow{2}{*}{Params}\\
\cmidrule(lr){2-3}
  & Train & Test &(M) \\
\midrule
RPT  \cite{Zhu2023FewShotMI} & 133.44 & 51.18 &45.1 \\
RobustEMD  \cite{Zhu2024RobustEMDDR} & 168.45 &89.32 &42.1\\
FAMNet  \cite{Bo2025FAMNetFM} & 86.32 & 22.72  &29.7\\
C-Graph  \cite{Bo2025ContrastiveGM} & 318.81  & 147.39 & 81.9\\
PSP \cite{Xu2026PSP} &132.53 &32.62 &63.8\\
\midrule
\textbf{GeoProto}&112.67 &12.87  &43.4\\
\bottomrule
\end{tabular}
\vspace{-2ex}
\end{table}

\paragraph{Computational efficiency.}
As shown in Table~\ref{tab:efficiency}, GeoProto remains lightweight: with 43.4M parameters, it is comparable to RobustEMD \cite{Zhu2024RobustEMDDR} (42.1M) while achieving substantially lower latency (112.67ms and 12.87ms for training and inference, respectively). This represents a $2.8{\times}$ reduction in training latency over C-Graph \cite{Bo2025ContrastiveGM} and a $1.8{\times}$ reduction in inference latency over the fastest competitor FAMNet \cite{Bo2025FAMNetFM}, demonstrating that geometry-aware enrichment via OSB and GAPE imposes negligible computational cost.

\begin{table*}[t]
\centering

\begin{minipage}{\linewidth}
\centering
\caption{Ablation on the number of bins $K$ (DSC \%). All models are trained on Abdominal CT. Best per column in \textbf{bold}.}
\vspace{-1ex}
\label{tab:ablation_K}
\resizebox{\linewidth}{!}{%
\begin{tabular}{lcccccccccccccc}
\toprule
\multirow{3}{*}{$K$} &
\multicolumn{5}{c}{\textit{Conventional FSMIS}} &
\multicolumn{5}{c}{\textit{Cross-modality}} &
\multicolumn{4}{c}{\textit{Cross-context}} \\
\cmidrule(lr){2-6}\cmidrule(lr){7-11}\cmidrule(lr){12-15}
& \multicolumn{5}{c}{Abdominal CT} & \multicolumn{5}{c}{Abdominal MRI} & CXR & \multicolumn{3}{c}{Cardiac b-SSFP} \\
\cmidrule(lr){2-6}\cmidrule(lr){7-11}\cmidrule(lr){13-15}
& Liver & LK & RK & Spleen & Mean
& Liver & LK & RK & Spleen & Mean
& Lung & BP & MYO & RV \\
\midrule
5  & 70.83  & 74.74  & 76.77 & 72.12 & 73.62
   & 61.52  & 74.04  & 74.64 & \textbf{68.76} & \textbf{69.74} & 83.93 & 67.38  & 39.60   & 58.81 \\
10 & 71.49 & 75.66 & \textbf{78.09} & 73.09       & 74.58 & \textbf{62.31} & 72.97    &\textbf{74.85} & 67.67  & 69.45
   & \textbf{84.50} & \textbf{71.26} &\textbf{46.75} & \textbf{65.09} \\
15 & 69.69 & \textbf{77.71} & 75.22 & 74.49       & 74.28 & 58.45 & \textbf{74.41} & 71.27       & 64.49 & 67.16 & 83.63  & 68.71 & 40.94       & 56.01 \\
20 & \textbf{71.52} & 76.47 & 76.60 &\textbf{76.93} & \textbf{75.38}
   & 60.81 & 72.57  & 73.89 & 67.50 & 68.69
   & 83.86 & 68.79  & 39.99 & 56.93 \\
\bottomrule
\end{tabular}}
\end{minipage}
\vspace{-2ex}
\end{table*}

\begin{table}[t]
\centering
\caption{Deterministic support-mask EDT \vs learned OSB (mean DSC \%). Best per column in \textbf{bold}.}
\vspace{-1ex}
\label{tab:det_edt}
\small
\setlength{\tabcolsep}{2.5pt}
\begin{tabular}{lccc}
\toprule
Setting & Deterministic EDT & Learned OSB (ours) & $\Delta$ \\
\midrule
CT$\to$MRI & 67.00 & \textbf{69.45} & $+2.45$ \\
MRI$\to$CT & 67.47 & \textbf{69.65} & $+2.18$ \\
LGE$\to$b-SSFP & 71.98 & \textbf{74.23} & $+2.25$ \\
b-SSFP$\to$LGE & 65.82 & \textbf{67.10} & $+1.28$ \\
\bottomrule
\end{tabular}
\vspace{-1ex}
\end{table}

\paragraph{Learned OSB \vs deterministic EDT.}
To answer a natural question -- whether our backbone learns geometric knowledge during training -- we replace OSB's learned bin prediction with a deterministic EDT bin embedding directly computed from the support mask. As shown in Table~\ref{tab:det_edt}, OSB is consistently better across all four cross-domain settings (+1.3 to +2.5 DSC), although both variants inject the same support mask. This suggests that the gains stem from learning the shape branch, not from the injection itself. A linear probe from encoded features to the ground-truth EDT bin (held-out foreground pixels) confirms this: the backbone trained with OSB is markedly more geometry-decodable than an otherwise identical one trained without it ($R^2{=}0.60$ \vs $0.49$), an effect a post-hoc EDT cannot produce.

\paragraph{Effect of bin number $K$.}
We select $K=10$ on the source-domain validation split and evaluate each variant on the target domain of each setting (Table~\ref{tab:ablation_K}; more details in Supp.~\ref{app:deploy}).
Performance is non-monotonic in $K$: $K{=}10$ is best on all cross-context targets and second-best in mean DSC on both abdominal datasets, where too few bins ($K{=}5$) under-quantise the interior topology and too many ($K{\geq}15$) introduce label sparsity that destabilises ordinal supervision. 
Nevertheless, the per-organ optimum varies (\eg, $K{=}15$ for LK and $K{=}20$ for spleen on Abdominal CT), suggesting that the optimal $K$ depends on organ-specific geometry such as structural complexity. This motivates adapting $K$ automatically during end-to-end training (discussed in Supp.~\ref{app:limitation}).

\begin{figure}[t]
  \centering
  \includegraphics[width=\linewidth]{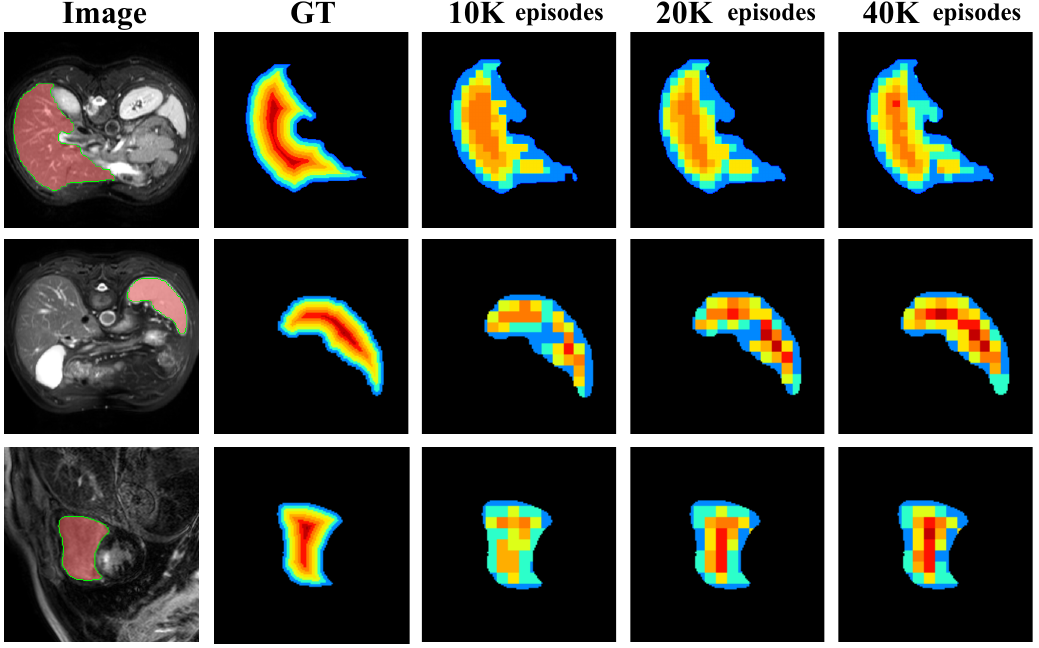}
  \vspace{-3.5ex}
  \caption{OSB Convergence Visualisation.}
  \label{fig:episode}
  \vspace{-1ex}
\end{figure}

\begin{figure}[t]
  \centering
  \includegraphics[width=\linewidth]{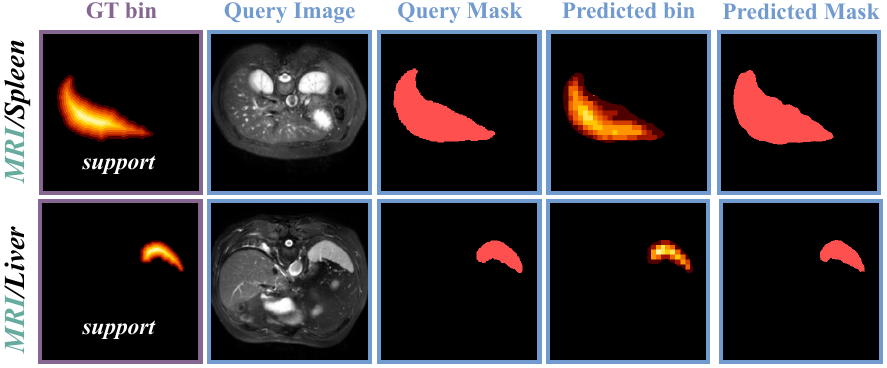}
  \vspace{-3.5ex}
  \caption{Qualitative results, with more samples in Figure~\ref{fig:qualitative}\rev{.}}
  \label{fig:vis}
  \vspace{-1.5ex}
\end{figure}

\section{Visualization}
\label{sec:vis}
Fig.~\ref{fig:episode} illustrates the evolution of predicted bin maps across training episodes on three representative organs. At 10K episodes, predictions are coarse and spatially inconsistent; by 40K episodes, the bin distributions converge to smooth, concentrically-structured patterns that vary monotonically from boundary to interior, closely matching the ground-truth EDT maps. This progressive convergence confirms that the ordinal training objective successfully instills geometric structure into the learned representations over the course of training.
Some qualitative results are shown in Fig.~\ref{fig:vis}, with more samples in Fig.~\ref{fig:qualitative}.

\section{Conclusion}
\label{sec:conclusion}

In this paper, we present GeoProto, a geometry-aware framework for the CD-FSMIS task. Rather than suppressing domain-specific cues, GeoProto exploits a stable anchor across imaging domains --- the ordinal boundary-to-centroid structure of anatomical organs. 
By learning this structure from standard binary masks and injecting it into local prototypes, GeoProto separates regions that look alike but occupy different anatomical strata.
Experiments across cross-modality, cross-sequence, and cross-context settings demonstrate state-of-the-art overall performance at low computational cost. 
More broadly, our results suggest that anatomical geometry is an under-exploited source of domain-invariant supervision, complementary to appearance-based domain alignment in medical image analysis.
Future work includes adapting the stratum granularity to organ-specific geometry and extending the ordinal strata to 3D volumetric context.

{
    \small
    \bibliographystyle{ieeenat_fullname}
    \bibliography{references}
}

\clearpage
\maketitlesupplementary
\appendix                          %

\setcounter{figure}{0}
\setcounter{table}{0}
\setcounter{equation}{0}
\renewcommand{\thetable}{S\arabic{table}}
\renewcommand{\thefigure}{S\arabic{figure}}
\renewcommand{\theequation}{S\arabic{equation}}
\renewcommand{\theHtable}{S\arabic{table}}
\renewcommand{\theHfigure}{S\arabic{figure}}
\renewcommand{\theHequation}{S\arabic{equation}}

\section{Preliminaries and Notation}
\label{app:prelim}

\noindent \textbf{Few-shot segmentation and the support/query setup.}
A few-shot segmentation model is trained on episodes sampled from base categories~\cite{Snell2017Prototypical,Wang2019PANetFI}. Each episode contains a \emph{support set} of annotated image--mask pairs and a \emph{query} image of the same category. The model learns to segment the query given the support. At test time the category is novel and only the support is annotated.

\noindent\textbf{Prototypical networks.} Prototypical networks~\cite{Snell2017Prototypical} represent each class by the prototypes, \ie, compact embeddings aggregated from labelled support features, and classify a query by its distance or similarity to these prototypes in a shared metric space. In few-shot segmentation~\cite{Wang2019PANetFI,Ouyang2022SelfSupervisedLF, Hansen2022AnomalyDF}, prototypes are typically obtained by masked average pooling of support features over the foreground and background regions, and each query pixel is labelled according to its similarity to them.

\noindent \textbf{Notation.}
Table~\ref{tab:notation} lists the symbols in the methodology section. Where a base letter is reused, we use distinct typeset forms for different kinds of mathematical objects. 

\begin{table}[h]
\centering
\caption{Notation used in Section~\ref{sec:method}.}
\vspace{-1ex}
\label{tab:notation}
\small
\begin{tabular}{ll}
\toprule
Symbol & Meaning \\
\midrule
\multicolumn{2}{l}{\textbf{\textit{Problem formulation}}} \\
$\mathcal{D}_s$, $\mathcal{D}_t$ & source / target domain \\
$\mathcal{C}_b$, $\mathcal{C}_n$ & base / novel category sets \\
$\mathcal{S}$, $\mathcal{Q}$ & support set, query set \\
$S$ & number of support shots \\
$I$, $M$ & image slice, binary mask ($H{\times}W$) \\
\midrule
\multicolumn{2}{l}{\textbf{\textit{Ordinal Shape Branch}}} \\
$F^s$, $F^q$ & support / query feature maps \\
$C$, $h{\times}w$ & feature channels, feature resolution \\
$\tilde{M}^s$, $\Omega$ & downsampled support mask, its foreground pixels \\
$K$, $\mathcal{K}$ & number of bins, bin set $\{0,\dots,K{-}1\}$ \\
$\mathcal{H}$ & OSB convolutional decoder \\
$\mathbf{Z}$, $\hat{\mathbf{Z}}$ & ground-truth / predicted bin map \\
$z(\mathbf{x})$, $\hat{z}(\mathbf{x})$ & ground-truth / predicted bin of pixel $\mathbf{x}$ \\
$\hat{p}_k(\mathbf{x})$ & predicted probability of bin $k$ at pixel $\mathbf{x}$ \\
$\beta$, $\lambda_{\mathrm{dist}}$ & foreground normalisation, ordinal penalty weight \\
\midrule
\multicolumn{2}{l}{\textbf{\textit{Geometry-Aware Prototype Enrichment}}} \\
$G$ & prototype grid size \\
$\mathcal{R}_i$ & spatial extent of cell $i$ \\
$\tau$, $n_{\mathrm{fg}}$ & occupancy threshold, number of retained cells \\
$\mathbf{p}^{\mathrm{fg}}_i$, $\mathbf{p}'_i$ & local / enriched foreground prototype \\
$\bar{d}_i$, $\mathbf{e}_i$ & expected bin of cell $i$, geometric offset \\
\midrule
\multicolumn{2}{l}{\textbf{\textit{Query prediction}}} \\
$\mathbf{p}^{\mathrm{bg}}_j$, $\mathbf{p}^{q}_i$ & background prototype, query prototype \\
$S_{\mathrm{fg}}$, $S_{\mathrm{bg}}$ & foreground / background score \\
$w_i$, $\alpha$ & attention weight, temperature \\
$\lambda_{\mathrm{geo}}$ & weight of $\mathcal{L}_{\mathrm{OSB}}$ in the total loss \\
\bottomrule
\end{tabular}
\vspace{-2ex}
\end{table}

\section{Bin Map Construction Details}
\label{app:bin_map}

\paragraph{Motivation.} The Ordinal Shape Branch (OSB) requires a per-pixel geometric label that captures each foreground pixel lies within the organ interior. A natural choice is the Euclidean distance-to-boundary transform (EDT), which assigns to each foreground pixel its shortest distance to the nearest background pixel. The EDT encodes the organ's interior topology in a continuous, geometry-faithful manner, which provides an ideal supervision signal for learning ordinal geometric representations.

\paragraph{Why discretise into bins?} Directly regressing the continuous EDT value is known to be unstable in deep networks, prone to outlier-driven gradient spikes and sensitive to organ scale. Instead, we reframe the problem as ordinal classification: the continuous EDT range is uniformly partitioned into K discrete bins, and the network is trained to predict which bin each foreground pixel belongs to. This formulation is more tractable, allows standard cross-entropy supervision, and naturally supports the ordinal penalty term $\mathcal{L}_\text{dist}$ (Eq. \ref{eq:dist} in the main paper) because adjacent bins have a well-defined ordinal relationship.

\paragraph{Formal construction.} Given the support mask downsampled to the feature resolution, $\tilde{M}^s$, the EDT is computed over all foreground pixels $\Omega = \{\mathbf{x} \mid \tilde{M}^s(\mathbf{x}) = 1\}$:
\begin{equation}
\mathrm{EDT}(\mathbf{x}) = \min_{\mathbf{x}' \notin \Omega} \|\mathbf{x} - \mathbf{x}'\|_2, \quad \mathbf{x} \in \Omega\rev{.}
\label{EDT1}
\end{equation}
The transform is then uniformly quantised into $K$ discrete bins to yield the ground-truth bin map $\mathbf{Z} \in \mathcal{K}^{h \times w}$, where $\mathcal{K} = \{0, 1, \ldots, K{-}1\}$:
\begin{equation}
z(\mathbf{x}) = \min\!\left(K{-}1,\ \left\lfloor \frac{\mathrm{EDT}(\mathbf{x})}{\max_{\mathbf{x}' \in \Omega} \mathrm{EDT}(\mathbf{x}')} \cdot K \right\rfloor \right)\rev{.}
\label{EDT2}
\end{equation}
Dividing by the per-slice maximum EDT value normalises the field to $[0, 1]$ before scaling, making the bin assignment invariant to absolute organ size, while the $\min$ assigns the deepest pixels to bin $K{-}1$. This is important because the same organ can appear at different physical scales across individuals and imaging resolutions.

\begin{table*}[h]
\centering
\caption{Mechanisms of the compared methods.}
\vspace{-1ex}
\label{tab:baselines}
\small
\begin{tabular}{llll}
\toprule
Method & Family & Mechanism & Intra-organ prior \\
\midrule
SSL-ALP~\cite{Ouyang2022SelfSupervisedLF} & FSMIS & local prototypes, superpixel self-supervision & no \\
ADNet~\cite{Hansen2022AnomalyDF} & FSMIS & anomaly-detection-inspired single prototype, supervoxels & no \\
RPT~\cite{Zhu2023FewShotMI} & FSMIS & region-enhanced prototypical transformer & no \\
GMRD~\cite{Cheng2024FewShotMI} & FSMIS & multiple generated representative descriptors & no \\
PATNet~\cite{Lei2022CrossDomainFS} & CD-FSS & transform features to a domain-agnostic space & no \\
PMNet~\cite{Chen2024PixelMN} & CD-FSS & pixel-to-pixel matching & no \\
IFA~\cite{Nie2024IFA} & CD-FSS & iterative support--query correspondence mining & no \\
APM-M~\cite{Tong2024LightweightFM} & CD-FSS & frequency masking & no \\
RobustEMD~\cite{Zhu2024RobustEMDDR} & CD-FSMIS & EMD matching suppressing texture-sensitive signal & no \\
FAMNet~\cite{Bo2025FAMNetFM} & CD-FSMIS & band-wise frequency de-biasing & no \\
C-Graph~\cite{Bo2025ContrastiveGM} & CD-FSMIS & cross-domain prototype graphs in a modality-agnostic space & no \\
DSM~\cite{zhu2025cross} & CD-FSMIS & support--query feature re-weighting, domain-robust channel selection &no \\
PSP~\cite{Xu2026PSP} & CD-FSMIS & polar position encoding, global shape descriptor as channel attention & no$^\ast$ \\
\rowcolor{gray!12} GeoProto (ours) & CD-FSMIS & geometry-aware prototype enrichment with ordinal strata & yes \\
\bottomrule
\end{tabular}
\par\vspace{3pt}
{
\footnotesize $^\ast$PSP uses image-level position and organ-level shape priors, but no intra-organ prior.\par}
\vspace{-1ex}
\end{table*}

\begin{table*}[h]
\centering
\caption{Comparison of geometric offset fusion modes (mean DSC \%). Best result per column in \textbf{bold}.}
\vspace{-1ex}
\label{tab:ablation_fusion}
\small
\setlength{\tabcolsep}{7pt}
\begin{tabular}{lccccccc}
\toprule
\multirow{2}{*}{Fusion mode} & \multicolumn{2}{c}{Cross-modality} & \multicolumn{2}{c}{Cross-sequence} & \multicolumn{3}{c}{Cross-context} \\
\cmidrule(lr){2-3}\cmidrule(lr){4-5}\cmidrule(lr){6-8}
 & CT$\to$MRI & MRI$\to$CT & LGE$\to$b-SSFP & b-SSFP$\to$LGE & CXR & Cardiac b-SSFP & MI-PMR \\
\midrule
\textbf{\emph{Additive}} (ours) & \textbf{69.45} & \textbf{69.65} & \textbf{74.23} & 67.10 & 84.50 & \textbf{61.03}& \textbf{61.24} \\
\textbf{\emph{Concat}}  & 63.93 & 68.57   &70.70   &67.63  & 82.70 & 51.99 & 51.11 \\
\textbf{\emph{Scale}} & 63.99 & 69.64  &72.02   &\textbf{70.94}  & \textbf{86.57} & 55.68 & 56.40 \\
\bottomrule
\end{tabular}
\end{table*}

\section{Failure Case Analysis}
\label{app:result}
\vspace{-0.5ex}
GeoProto's geometric enrichment relies on the boundary-to-centroid ordinal structure being informative, \ie, on foreground pixels being spread across multiple strata so that local prototypes from different regions receive distinct offsets. Fig.~\ref{fig:geo_prior} reports the $K$-bin EDT distribution, where compact organs such as RK and spleen exhibit relatively flat distributions, whereas the liver is strongly skewed towards the boundary: $59.6\%$ of its pixels fall in bin $0$ and only $0.4\%$ near the centroid. Because the EDT is normalised by the maximum interior distance of each slice, a few deep pixels in the large, multi-lobed liver cross-section compress most of the parenchyma into the lowest strata. Consequently, most local cells obtain similar expected strata $\bar{d}_i$ and hence similar offsets $\mathbf{e}_i$, which limits GAPE's ability to differentiate prototypes by their position within the organ, leaving matching on the liver dominated by appearance cues.
This analysis is consistent with the quantitative results in Table~\ref{tab:table1}. The liver is the organ on which GeoProto trails the best method most ($62.31\%$ vs.\ $73.01\%$ under CT$\to$MRI and $69.65\%$ vs.\ $77.69\%$ under MRI$\to$CT), whereas on the more compact kidneys and spleen it achieves the best or near-best results.

\section{Comparison with Baselines}
\label{app:baselines}

Table~\ref{tab:baselines} summarises the mechanism each compared method relies on and what it leaves unaddressed relative to GeoProto. Most of them attack domain shift from an appearance-centric angle, while PSP~\cite{Xu2026PSP} additionally exploits image-level position and organ-level shape cues. However, none provides an explicit intra-organ geometric axis along which prototypes are matched.

In our main experiments (\cref{tab:table1,tab:table2,tab:table7_extended}), all reported results use a single backbone (ResNet-101, MS-COCO initialised, $32{\times}32$ feature map) and a single fixed $G{=}8$, $K{=}10$ across all datasets, organs and settings.

\begin{table*}[h]
\centering
\caption{Analysis of background prototype geometric enrichment (mean DSC \%). Best result per column in \textbf{bold}.}
\vspace{-1ex}
\label{tab:ablation_bg}
\small
\begin{tabular}{lccccc}
\toprule
\multirow{2}{*}{$+$BG-Enrich} & Cross-modality & Cross-sequence & \multicolumn{3}{c}{Cross-context} \\
\cmidrule(lr){2-2} \cmidrule(lr){3-3} \cmidrule(lr){4-6}
 & CT$\to$MRI & b-SSFP $\to$ LGE & Chest X-ray  & ISIC & Cardiac b-SSFP\\
\midrule
None (ours)        & \textbf{69.45} & \textbf{67.10} & \textbf{84.50} &\textbf{47.45} & \textbf{61.03}\\
Dist-to-FG embed   & 64.40 & 56.47 & 71.88 &44.18 & 34.80  \\
\bottomrule
\end{tabular}
\end{table*}

\begin{table*}[h]
\centering
\caption{Analysis of query-side geometric re-weighting (mean DSC \%). Best result per column in \textbf{bold}.}
\vspace{-1ex}
\label{tab:ablation_query}
\small
\begin{tabular}{lccccc}
\toprule
\multirow{2}{*}{Query re-weighting} & Cross-modality & Cross-sequence & \multicolumn{3}{c}{Cross-context} \\
\cmidrule(lr){2-2} \cmidrule(lr){3-3} \cmidrule(lr){4-6}
 & CT$\to$MRI & b-SSFP $\to$ LGE & Chest X-ray  & ISIC & Cardiac b-SSFP\\
\midrule
None (ours)         & 69.45 &67.10 & 84.50 &47.45 & \textbf{61.03}  \\
$\sigma = 0.1$        & 69.45 &\textbf{69.61} & 82.55 & 42.84 &56.56  \\
$\sigma = 0.3$        & \textbf{70.48} &69.43 &81.30 &39.50 &60.91  \\
$\sigma = 0.5$        & 69.54 &67.76 &\textbf{86.89} &\textbf{50.20} &57.52 \\
\bottomrule
\end{tabular}
\end{table*}

\section{Ablation Studies}
\label{app:ablation}

\subsection{Geometric Offset Fusion Mode}
\label{app:fusion}
We compare three fusion modes for injecting geometric embedding $\mathbf{e}_i$ into each local prototype: \textbf{\emph{concat}}, which concatenates $\mathbf{e}_i$ to $\mathbf{p}^{\mathrm{fg}}_i$ and projects back to the original dimension via a linear layer, and \textbf{\emph{scale}}, which applies a multiplicative gate $\mathbf{p}'_i = \mathbf{p}^{\mathrm{fg}}_i \odot (1 + \mathrm{softplus}(\mathbf{e}_i))$.
Including the \textbf{\emph{additive}} fusion we used in Eq.~\ref{eq:enriched}, all three variants share the same MLP architecture and are initialised identically.

As shown in Table~\ref{tab:ablation_fusion}, our additive fusion achieves the best overall performance. Scale gate outperforms additive fusion on b-SSFP$\to$LGE and CXR, suggesting that multiplicative modulation can be more effective when appearance and geometric cues are well-aligned. Concatenation with projection performs poorly across most settings, likely because the additional linear projection introduces parameters that are difficult to optimise under episodic training with limited support, and the concatenation disrupts the near-zero initialisation strategy that allows geometric influence to accrue gradually.

\subsection{Differentiable Ordinal Weight}
\label{app:diff_ord}

The weighting term $\omega=|\hat{z}(\mathbf{x})-z(\mathbf{x})|/K$ in $\mathcal{L}_\text{dist}$ (Eq.~\ref{eq:dist}) uses the arg-max predicted bin and is not differentiable with respect to the logits, although gradients flow through the $-\log(1-\max_k\hat{p}_k(\mathbf{x}))$ factor.
Therefore, we implement a fully differentiable variant that replaces it with the expected bin, $\omega_{\text{diff}}=\bigl|\sum_k k\,\hat{p}_k(\mathbf{x})-z(\mathbf{x})\bigr|/K$, a smooth function of the whole softmax distribution.
The two variants perform comparably, yielding mean DSC of 69.45\% \vs 68.46\% on CT$\to$MRI and 69.65\% \vs 70.08\% on MRI$\to$CT for $\omega$ and $\omega_{\text{diff}}$, respectively.

\subsection{Analysis: Whether to use background?}
\label{app:bg_sdf}

By design, GAPE enriches foreground prototypes only, as the EDT is defined exclusively on foreground pixels. A natural question is whether enriching background prototypes, using the distance to the nearest foreground boundary as a spatial cue for each background cell, provides additional benefit. We implement this variant (BG-Enrich) and compare it against the default in Table~\ref{tab:ablation_bg}.

In Sec.~\ref{sec:method}, we hypothesise that 
\textit{``the geometric stratum is undefined for background pixels''}
because background regions are heterogeneous across episodes: they encompass non-organ tissue whose spatial relationship to the foreground varies widely, so a single distance-to-boundary signal does not constitute a reliable structural prior for the background. The experimental results in Table~\ref{tab:ablation_bg} confirm this: background enrichment yields substantial degradation across settings, justifying the design choice of leaving background prototypes unenriched.

\subsection{Analysis: Query-Side Geometric Re-weighting}
\label{app:query_sdf}

Our default design applies geometric enrichment exclusively on the support side. A plausible extension is to exploit the query's own predicted bin map to re-weight prototype matching at inference: query pixels predicted to lie near the organ boundary should preferentially match boundary-region prototypes, and conversely for centroid-region pixels.

Applying OSB to the query features yields bin probabilities $\hat{p}^{q}_k(\mathbf{x})$, from which the normalised expected stratum of query pixel $\mathbf{x}$ is
\begin{equation}
\hat{g}(\mathbf{x}) = \frac{1}{K-1}\sum_{k=0}^{K-1} k\cdot\hat{p}^{q}_k(\mathbf{x}),
\end{equation}
and the normalised stratum of prototype $i$ is $\bar{g}_i = \bar{d}_i/(K-1)$, with $\bar{d}_i$ from Eq.~\ref{eq:avgbin}.
A soft affinity gate then measures their geometric consistency:
\begin{equation}
\gamma_i(\mathbf{x}) = \exp\!\Bigl(-\frac{|\hat{g}(\mathbf{x}) - \bar{g}_i|}{\sigma}\Bigr),
\end{equation}
where $\sigma$ controls the gate sharpness. Finally, the gate scales each cosine similarity before the softmax that defines the attention weights in Sec.~\ref{sec:classification}:
\begin{equation}
w_i(\mathbf{x}) = \mathrm{softmax}_i\bigl(\alpha \cos(F^q(\mathbf{x}), \mathbf{p}_i')\cdot\gamma_i(\mathbf{x})\bigr).
\end{equation}
Prototypes whose geometric stratum is inconsistent with the query pixel are thus down-weighted during attention competition.
Table~\ref{tab:ablation_query} reports results for $\sigma$ $\in \{0.1, 0.3, 0.5\}$ against the default.
No single value of $\sigma$ achieves consistent gains across all evaluation settings, revealing a sensitivity to $\sigma$ that varies with domain and anatomy. We attribute this to the fact that the OSB is trained exclusively on source-domain features, so the query bin map predicted on novel-domain features carries higher uncertainty than the support's supervised prediction. Restricting enrichment to the masked support side is therefore the more robust design.

\begin{table*}[h]
\centering
\caption{GeoProto over 5 random seeds on both cross-modality directions (DSC \%).}
\vspace{-1ex}
\label{tab:seeds}
\small
\begin{tabular}{llccccc}
\toprule
Setting & & Liver & RK & LK & Spleen & Overall \\
\midrule
\multirow{2}{*}{CT$\to$MRI} & mean $\pm$ std & 63.66 $\pm$ 1.13 & 74.92 $\pm$ 1.15 & 72.38 $\pm$ 1.00 & 66.94 $\pm$ 0.77 & 69.47 $\pm$ 0.65 \\
 & 95\% CI & [62.26, 65.07] & [73.50, 76.35] & [71.15, 73.62] & [65.99, 67.90] & [68.67, 70.30] \\
\midrule
\multirow{2}{*}{MRI$\to$CT} & mean $\pm$ std & 70.11 $\pm$ 0.62 & 68.90 $\pm$ 2.75 & 68.95 $\pm$ 2.15 & 69.41 $\pm$ 1.71 & 69.34 $\pm$ 0.75 \\
 & 95\% CI & [69.33, 70.88] & [65.48, 72.31] & [66.28, 71.62] & [67.29, 71.54] & [68.41, 70.27] \\
\bottomrule
\end{tabular}
\end{table*}

\section{Statistical and Sensitivity Analyses}
\label{app:statsens}

\subsection{Multi-seed Statistics}
\label{app:stats}

We retrain GeoProto with 5 random seeds on both cross-modality directions (CT$\to$MRI and MRI$\to$CT) and report mean $\pm$ std and a 95\% confidence interval (Student-$t$, df $=4$) in Table~\ref{tab:seeds}. The overall std of $0.65$ (CT$\to$MRI) and $0.75$ (MRI$\to$CT) shows that the reported performance is stable across training randomness.

\subsection{Sensitivity to Support-slice Selection}
\label{app:support}

The 1-shot protocol includes substantial support-slice variability across episodes, as support slices are sampled from different positions along the through-plane axis, ranging from the superior/inferior extremities of the organ volume to its mid-section.

To test sensitivity to support-slice selection, we further evaluate GeoProto on CT$\to$MRI setting. For each query slice, we draw the support at three SSIM levels around the realistic operating band (P25/P50/P75 of all query--candidate pairs), alongside the default protocol, and deliberately avoid the extremes: near-maximal SSIM is oracle-like and near-minimal SSIM is unrepresentative of practical use.
As shown in Table~\ref{tab:support}, DSC varies by 7.8 points across this band and degrades gracefully rather than collapsing, while the compact kidneys remain markedly stable, which is consistent with our geometric prior.

\begin{table}[t]
\centering
\caption{Sensitivity to the support slice on CT$\to$MRI (DSC \%).}
\vspace{-1ex}
\label{tab:support}
\small
\setlength{\tabcolsep}{3pt}
\begin{tabular}{lcccccc}
\toprule
Support & SSIM & Mean & Liver & RK & LK & Spleen \\
\midrule
High & 0.570 & 72.9 & 69.6 & 74.1 & 75.1 & 72.7 \\
\rowcolor{gray!12} Default (mid) & 0.550 & 69.5 & 62.3 & 74.8 & 73.0 & 67.7 \\
Moderate & 0.523 & 69.0 & 61.6 & 72.1 & 72.2 & 70.0 \\
Low & 0.498 & 65.1 & 57.2 & 68.6 & 70.7 & 63.8 \\
\bottomrule
\end{tabular}
\end{table}

\subsection{Multi-shot Evaluation}
\label{app:kshot}
Our main protocol follows prior CD-FSMIS work in adopting the 1-way 1-shot setting. To examine whether GeoProto benefits from additional support, we evaluate the same trained model under a 5-shot protocol (five evenly spaced slices along the through-plane axis of the support volume) on both cross-modality directions.
As shown in Table~\ref{tab:kshot}, performance improves consistently across all organs as the number of shots increases, \eg, mean DSC rises from 69.45\% to 74.31\% on CT$\to$MRI and from 69.65\% to 76.72\% on MRI$\to$CT. Since the model is not retrained, this indicates that GeoProto's prototype matching effectively aggregates geometric and appearance cues from additional support slices at inference time, without requiring any architectural change.

\begin{table}[t]
\centering
\caption{1-shot \vs 5-shot evaluation (DSC \%).}
\vspace{-1ex}
\label{tab:kshot}
\small
\setlength{\tabcolsep}{3pt}
\begin{tabular}{llccccc}
\toprule
Setting & Support & Liver & RK & LK & Spleen & Mean \\
\midrule
\multirow{3}{*}{CT$\to$MRI} & 1-shot & 62.3 & 74.8 & 73.0 & 67.7 & 69.45 \\
 & 5-shot & 72.5 & 76.1 & 74.7 & 73.9 & 74.31 \\
 & $\Delta$ & $+10.2$ & $+1.3$ & $+1.7$ & $+6.2$ & $+4.86$ \\
\midrule
\multirow{3}{*}{MRI$\to$CT} & 1-shot & 69.6 & 69.7 & 71.3 & 68.0 & 69.65 \\
 & 5-shot & 80.3 & 74.6 & 76.5 & 75.5 & 76.72 \\
 & $\Delta$ & $+10.7$ & $+4.9$ & $+5.2$ & $+7.5$ & $+7.08$ \\
\bottomrule
\end{tabular}
\end{table}

\begin{table}[t]
    \centering
    \caption{Source-domain results (DSC\%) on conventional FSMIS. 
    Best results per column in \textbf{bold}.}
    \label{tab:source_domain_results}
    \vspace{-1ex}
    \footnotesize
    \addtolength{\tabcolsep}{-3pt}
    \newcolumntype{Y}{>{\centering\arraybackslash}X}
    \small
    \begin{tabularx}{\linewidth}{l YYYYY}
    \toprule
    \multirow{2}{*}{Method} & \multicolumn{5}{c}{Abdominal CT} \\
    \cmidrule(lr){2-6}
    & Liver & LK & RK & Spleen & Mean \\
    \midrule
    SSL-ALP \cite{Ouyang2022SelfSupervisedLF}   & 78.29 & 72.36 & 71.81 & 70.96 & 73.35 \\
    ADNet \cite{Hansen2022AnomalyDF}   & 77.24 & 72.13 & \textbf{79.06} & 63.48 & 72.97 \\
    RobustEMD \cite{Zhu2024RobustEMDDR} & \textbf{79.30} & 66.67 & 54.75 & 67.10 & 66.96  \\
    FAMNet \cite{Bo2025FAMNetFM}   & 74.29 & 71.14 & 66.13 & 70.08 & 70.41  \\
    \midrule
    Ours & 71.49 & \textbf{75.66} & 78.09 & \textbf{73.09} & \textbf{74.58} \\
    \bottomrule
    \end{tabularx}
\end{table}

\subsection{Source-domain Performance}
\label{app:source_domain}

We also evaluate GeoProto on the conventional FSMIS setting, where both training and evaluation are conducted within the same imaging domain, reported in Table~\ref{tab:source_domain_results}. 
On Abdominal CT, GeoProto reaches a mean DSC of $74.58\%$ and, as in the cross-domain results, performs well on the compact kidneys and spleen.
This strong in-domain performance further demonstrates that the geometric prior introduced by GAPE is complementary to appearance-based matching rather than conflicting with it: even when domain shift is absent, enriching prototypes with ordinal geometric offsets provides additional discriminative structure that improves segmentation accuracy within the source domain.

\begin{table*}[h]
\centering
\caption{Effect of prototype grid size $G$ (mean DSC \%). Best result per column in \textbf{bold}.}
\vspace{-1ex}
\label{tab:ablation_grid}
\small
\setlength{\tabcolsep}{7pt}
\begin{tabular}{lccccccc}
\toprule
\multirow{2}{*}{$G$} & \multicolumn{2}{c}{Cross-modality} & \multicolumn{2}{c}{Cross-sequence} & \multicolumn{3}{c}{Cross-context} \\
\cmidrule(lr){2-3}\cmidrule(lr){4-5}\cmidrule(lr){6-8}
 & CT$\to$MRI & MRI$\to$CT & LGE$\to$b-SSFP & b-SSFP$\to$LGE & CXR & Cardiac b-SSFP & MI-PMR \\
\midrule
4  & \textbf{69.58} &68.19  & 69.88 &65.99 & \textbf{85.14} & 60.35 & 56.69 \\
\rowcolor{gray!12} 8  & 69.45 &\textbf{69.65} & \textbf{74.23} & \textbf{67.10} & 84.50 & \textbf{61.03} & \textbf{61.24}\\
16 & 66.29 &64.43 & 65.40 & 66.73 & 82.00 & 55.62 & 55.85 \\
32 & 64.11 &61.62  &65.34 &65.21 & 84.66 & 56.65 & 56.29 \\
\bottomrule
\end{tabular}
\end{table*}

\section{Hyperparameter Analysis}
\label{app:deploy}

Both hyperparameters are selected on the source-domain validation split following DomainBed~\cite{gulrajani2020search} and kept fixed across all datasets, organs and settings ($K{=}10$, $G{=}8$). They differ in how they depend on the input. The bin number $K$ discretises the per-slice normalised distance-to-boundary transform (Eq.~\ref{EDT2}), so it indexes relative boundary-to-centroid position, decoupled from both organ size and feature resolution, and is expected to transfer across backbones unchanged. In contrast, the prototype grid size $G$ is spatial: on a backbone with a different feature resolution, $G$ should be scaled to preserve the cell coverage of $4{\times}4$ feature pixels.

\subsection{Effect of Prototype Grid Size $G$}
\label{app:grid_size}

The prototype grid size $G$ controls the spatial granularity of local prototype pooling: a larger $G$ yields finer-grained prototypes but increases the risk of pooling too few foreground pixels per cell, while a smaller $G$ sacrifices spatial resolution.
We ablate $G \in \{4, 8, 16, 32\}$ on the target domain of each setting to assess sensitivity, with all other hyperparameters fixed to their defaults (Table~\ref{tab:ablation_grid}).

The results reveal a consistent optimum at $G=8$ across most evaluation settings. At $G{=}4$, the coarser partition merges anatomically distinct regions into the same prototype cell, reducing the geometric discriminability that GAPE relies on to differentiate boundary-concentrated from centroid-concentrated prototypes. At $G{=}16$ and beyond, each cell covers only a few feature pixels (a single one at $G{=}32$), so local prototypes and their expected strata $\bar{d}_i$ are estimated from very few samples. The resulting prototypes are noisy and close to pixel-level matching, which makes them sensitive to appearance variation under domain shift. CXR is less sensitive to $G$, as the lungs are large, simple shapes whose strata remain well estimated even at fine or coarse grids.

\subsection{Test-Time Adaptation}
\label{app:ttft}
As a practical adaptation recipe, we explore lightweight test-time fine-tuning: a few gradient steps on the support pair's own annotation. We test it on WBC~\cite{Zheng2018WBC}, a microscopic white-blood-cell dataset of 200 images with two categories (cytoplasm and nucleus) that is never used in the paper, with the same Abdominal-CT-trained model.
As shown in Table~\ref{tab:ttft}, the model already transfers to this out-of-modality setting, and lightweight fine-tuning yields further gains, indicating that GeoProto can directly employ test-time adaptation during inference.

\begin{table}[t]
\centering
\caption{Lightweight test-time fine-tuning on the support pair (mean DSC \%). Model trained on Abdominal CT.}
\vspace{-1ex}
\label{tab:ttft}
\small
\setlength{\tabcolsep}{3pt}
\begin{tabular}{lccc}
\toprule
Target & GeoProto & + fine-tuning & $\Delta$ \\
\midrule
Abdominal MRI & 69.45 & 74.21 & $+4.76$ \\
WBC (unseen dataset) & 63.20 & 69.66 & $+6.46$ \\
\bottomrule
\end{tabular}
\end{table}

\section{Limitation}
\label{app:limitation}

\noindent\textbf{Morphology-dependent gains.} The boundary-to-centroid ordinal prior is most effective for compact, roughly convex organs. For large homogeneous organs such as the liver, adjacent cells receive similar offsets, while for irregular targets such as skin lesions, the interior is not concentric. GeoProto's gains are correspondingly limited for these targets.

\vspace{1ex}

\noindent \textbf{A fixed bin number.} We use a single $K{=}10$ for every organ and setting, although the optimal $K$ varies across organs, 
as the number of strata a structure can support depends on its shape complexity. A natural extension is to adapt $K$ to each target automatically during end-to-end training, for instance by learning the bin edges jointly with the OSB.

\vspace{1ex}

\noindent \textbf{2D slice-based processing.} In CD-FSMIS task, GeoProto operates on 2D image slices and does not exploit the volumetric continuity inherent in 3D medical imaging. However, the EDT structure of an organ is geometrically more informative in 3D space. Extending our model to volumetric 3D processing, with 3D geometric supervision, represents a natural direction for our future work.

\vspace{1ex}

\noindent\textbf{Gap in practical application.} Our geometric prior implicitly assumes that target structures have well-defined boundaries and largely preserved gross morphology, so that their boundary-to-centroid organisation is consistent across individuals. As mentioned in Sec.~\ref{sec:intro}, in practice, this assumption is often violated for extreme pathological conditions, \eg tumours, where reliable segmentation is often most needed. Future work may explore dynamic topological encodings that adapt the geometric representation to the shape complexity of each target.

\begin{figure*}[h]
  \centering
  \includegraphics[width=\linewidth]{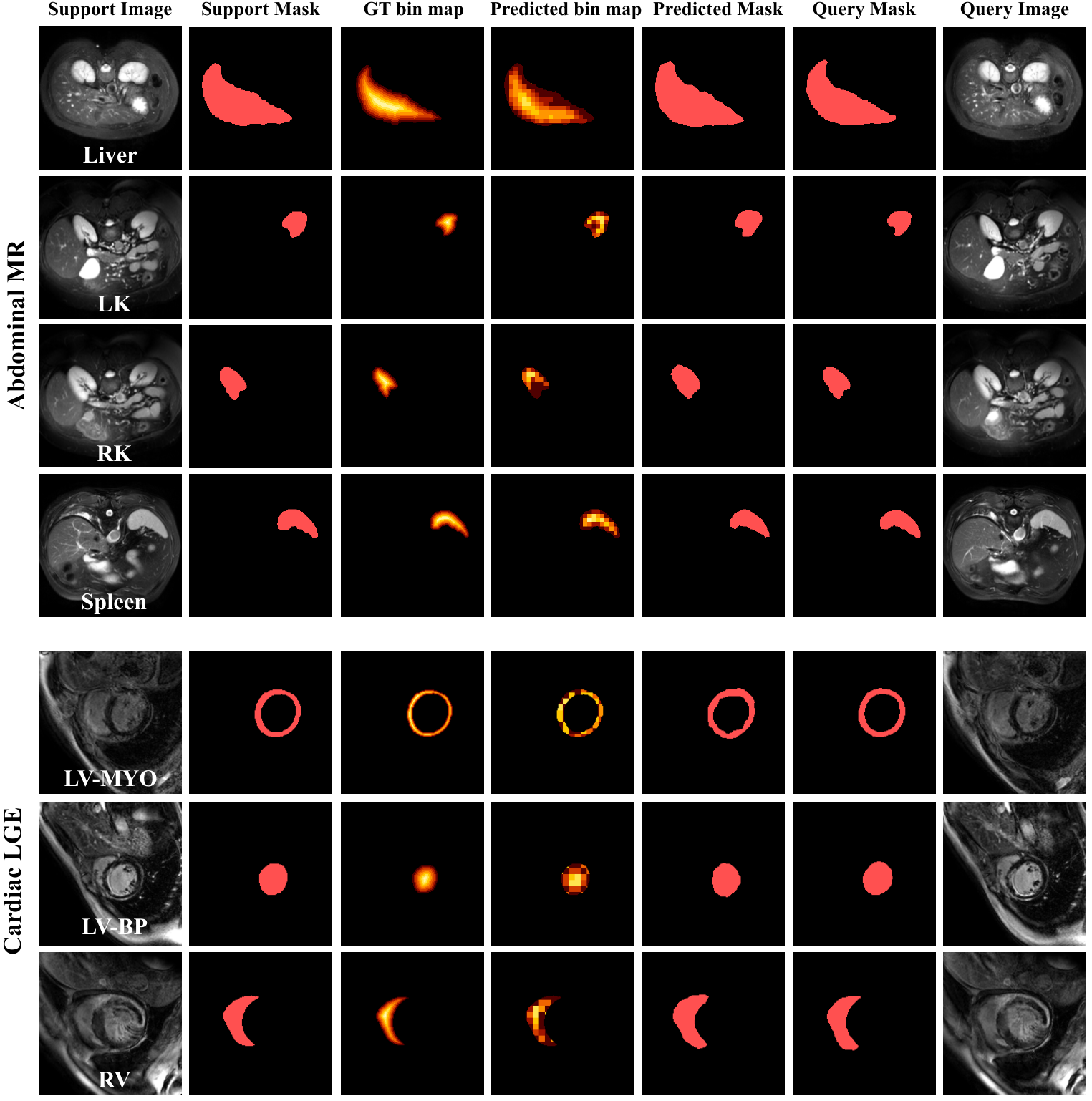}
  \caption{Qualitative results. From left to right: support image, support mask, ground-truth and predicted bin maps of the support, predicted query mask, ground-truth query mask, and query image.}
  \label{fig:qualitative}
\end{figure*}

\end{document}